\documentclass{article}
\usepackage{spconf,amsmath,graphicx}
\usepackage{xcolor}
\usepackage{algorithm}
\usepackage{algorithmic}
\usepackage{amssymb}

\newlength\myindent
\newcommand\bindent{%
  \begingroup
  \setlength{\itemindent}{\myindent}
  \addtolength{\algorithmicindent}{\myindent}
}
\newcommand\eindent{\endgroup}

\title{GMAR: Gradient-Driven Multi-Head Attention Rollout for Vision Transformer Interpretability}
\name{Sehyeong Jo$^{1}$ \qquad Gangjae Jang$^{1}$ \qquad Haesol Park$^{2,\star}$\thanks{$^{\star}$Corresponding author: haesol@kist.re.kr}}
  
\address{$^{1}$University of Colorado Boulder, Boulder, CO, USA \\
  $^{2}$Korea Institute of Science and Technology(KIST), Seoul, Korea}

\begin{document}
%
\maketitle
\begin{abstract}
The Vision Transformer (ViT) has made significant advancements in computer vision, utilizing self-attention mechanisms to achieve state-of-the-art performance across various tasks, including image classification, object detection, and segmentation. Its architectural flexibility and capabilities have made it a preferred choice among researchers and practitioners. However, the intricate multi-head attention mechanism of ViT presents significant challenges to interpretability, as the underlying prediction process remains opaque. A critical limitation arises from an observation commonly noted in transformer architectures: "Not all attention heads are equally meaningful." Overlooking the relative importance of specific heads highlights the limitations of existing interpretability methods. To address these challenges, we introduce Gradient-Driven Multi-Head Attention Rollout (GMAR), a novel method that quantifies the importance of each attention head using gradient-based scores. These scores are normalized to derive a weighted aggregate attention score, effectively capturing the relative contributions of individual heads. GMAR clarifies the role of each head in the prediction process, enabling more precise interpretability at the head level. Experimental results demonstrate that GMAR consistently outperforms traditional attention rollout techniques. This work provides a practical contribution to transformer-based architectures, establishing a robust framework for enhancing the interpretability of Vision Transformer models.
\end{abstract}
\section{Introduction}
\label{sec:intro}

The Vision Transformer (ViT) has emerged as a transformative architecture in computer vision~\cite{dosovitskiy2020image}, achieving state-of-the-art performance in tasks like image classification and object detection. Unlike traditional convolutional neural networks (CNNs) that rely on localized feature extraction, ViTs leverage a global perspective through the self-attention mechanism. As ViTs gain traction in critical domains such as medical imaging and autonomous systems, ensuring their interpretability is vital for reliability, transparency, and trust in real-world applications.

Despite their success, ViTs face significant interpretability challenges due to the complexity of their multi-head attention mechanism, which captures diverse aspects of input data. Existing explainable AI (XAI) methods adapted from CNNs~\cite{ali2022xai} often fail to address the unique characteristics of attention-based architectures. Attention rollout~\cite{abnar2020quantifying} techniques provide insights into the contribution of attention layers but overlook the varying importance of individual attention heads, potentially leading to incomplete or misleading interpretations. This highlights the need for refined interpretability methods designed specifically for Vision Transformers.

\begin{figure*}
    \centering
    \includegraphics[width=\textwidth]{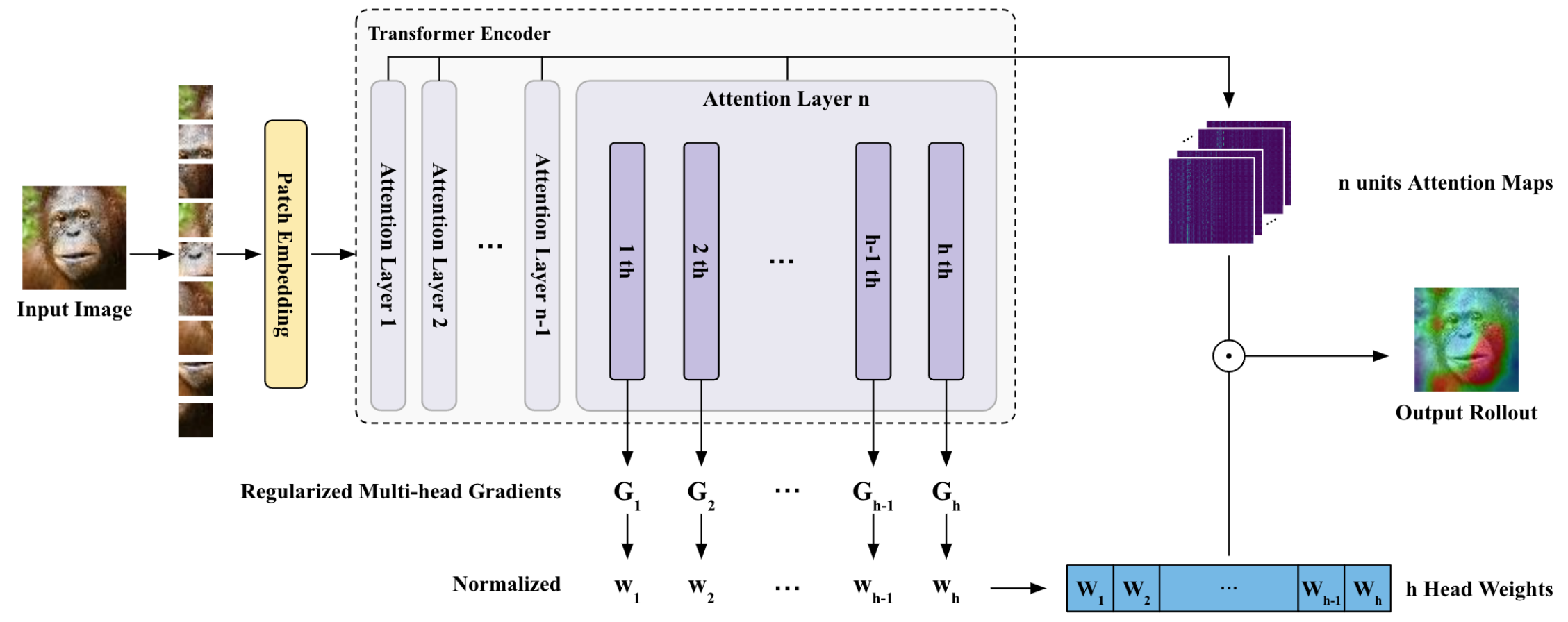}
    \caption{Overview of the Gradient-Driven Multi-Head Attention Rollout (GMAR) algorithm structure as applied to Vision Transformers, detailing the computation of head weights in multi-head gradients and illustrating the attention mechanism across layers and heads to enhance interpretability.}
    \label{fig:overview}
\end{figure*}

To address these challenges, we introduce Gradient-Driven Multi-Head Attention Rollout (GMAR), a novel interpretability framework specifically designed for Vision Transformers (Figure~\ref{fig:overview}). GMAR leverages class-specific gradient information to quantify the contribution of individual attention heads to the model’s predictions. Gradient scores are calculated for each head and normalized using L1 and L2 norms, ensuring a consistent and interpretable comparison of head importance. These normalized scores are then used as weights applied directly to the attention layers. The weighted attention layers are subsequently rolled out across the network, generating a comprehensive and interpretable attention map that highlights how individual heads and attention layers contribute to the model's predictions. Gradient-based XAI methods effectively capture regions that contribute most significantly to the prediction of a specific class by directly or indirectly utilizing class-specific gradients. Meanwhile, attention rollout techniques excel in interpreting the internal self-attention mechanisms of Transformers but may not accurately reflect the actual contribution of attention heads, as they often incorporate dispersed or contextual information rather than class-specific sensitivity. By combining the strengths of gradient-based methods in reflecting class-specific contributions and attention rollout’s ability to elucidate the self-attention mechanism, GMAR offers a refined interpretability framework.

\section{RELATED WORK}
\label{sec:related}

\subsection{Explainable AI}
\label{ssec:re_xai}

The increasing complexity of deep learning models highlights the need for reliability and transparency, driving advancements in XAI. Model-agnostic techniques, such as LIME(Local Interpretable Model-agnostic Explanations)~\cite{ribeiro2016should}, utilize local surrogate models to explain predictions, while SHAP(SHapley Additive exPlanations)~\cite{scott2017unified} employs game-theoretic principles to assess feature importance. Gradient-based methods, including Grad-CAM(Gradient-weighted Class Activation Mapping)~\cite{selvaraju2017grad}, generate class-specific activation maps from the final CNNs, whereas LRP(layer-wise relevance propagation)~\cite{bach2015pixel} retraces relevance through network layers to elucidate model predictions.

XAI methods developed specifically for transformer models include Attention Rollout and Attention Flow~\cite{abnar2020quantifying}. They represent key XAI approaches for transformer models, with Attention Rollout visualizing input significance by integrating attention weights across layers and Attention Flow revealing decision-making pathways by mapping information flow between attention heads. For ViTs, techniques like ViT-CX~\cite{xie2023vit} leverage the semantic richness of embeddings, and approaches such as FovEx~\cite{panda2024human}, inspired by human visual systems, emphasize the specialized requirements of explaining transformer-based models. 

\subsection{Vision Transformer}
\label{ssec:re_vis}

The transformer architecture~\cite{vaswani2017attention} has had a profound impact, demonstrating its increasing significance across various domains. Transformers, built on an encoder-decoder structure, leverage the self-attention mechanism to capture global context in input sequences, improving performance. Their parallel processing capability further enhances computational efficiency and scalability. Prominent encoder-based models such as BERT~\cite{devlin2018bert} and RoBERTa~\cite{liu2019roberta}, along with decoder-focused architectures like GPT~\cite{brown2020language} and LLaMA~\cite{touvron2023llama}, demonstrate the versatility of Transformer-based approaches. Beyond NLP, these models have been increasingly adopted in other fields such as computer vision, further showcasing the adaptability of this architecture.

ViTs have emerged as a prominent approach in computer vision by adapting the transformer architecture to image data. By segmenting input images into fixed-size patches and processing them through a transformer encoder, ViTs leverage self-attention mechanisms to learn global relationships, offering a distinct advantage over CNNs. Variants of ViTs, such as DeiT (Data-efficient Image Transformers)~\cite{touvron2021training} utilizing knowledge distillation, BEiT (BERT pre-training of Image Transformers)~\cite{bao2021beit} incorporating masked image modeling and VQ-VAE-based self-supervised learning, and Swin Transformer~\cite{liu2021swin} with a hierarchical structure and shifted window self-attention, have advanced the field significantly.

\section{Proposed Method}
\label{sec:method}

\subsection{Overview of Attention Rollout}
\label{ssec:me_ar}

Attention rollout is a technique designed to interpret and visualize the internal mechanisms of ViTs. It aggregates attention weights across multiple layers to quantify the influence of input tokens on the model’s final prediction. By recursively combining attention maps from all layers, attention rollout provides a holistic view of the information flow within the model, thereby enhancing interpretability. 

Formally, attention rollout computes the cumulative attention as:
\begin{equation}
    A_{\text{rollout}} = \prod_{l=1}^{L} \left( \mathbf{A}^{(l)} + \mathbb{I} \right),
\end{equation}

where \( \mathbf{A}^{(l)} \in \mathbb{R}^{N \times N} \) denotes the normalized attention matrix at layer \( l \), \( \mathbb{I} \in \mathbb{R}^{N \times N} \) is the identity matrix accounting for residual connections, and \( L \) is the total number of layers. This recursive aggregation ensures that the contributions from all layers are captured effectively, yielding a comprehensive attention map.

\begin{figure}
    \centering
    \includegraphics[width=\columnwidth]{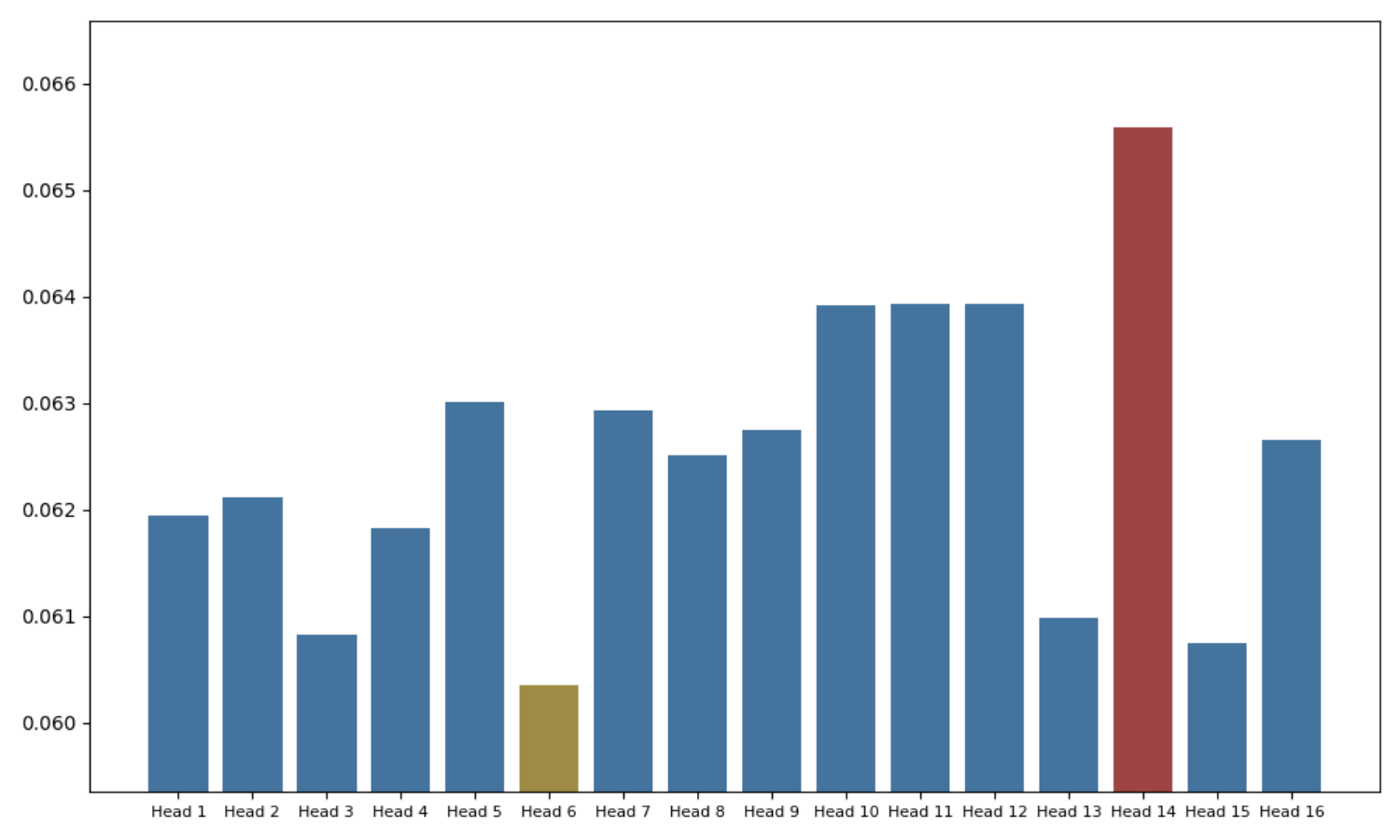}
    \caption{Comparing attention head weights, with the highest and lowest values highlighted in red and yellow}
    \label{fig:headbar}
\end{figure}

\subsection{Defining Head Importance}
\label{ssec:me_head}

In the attention rollout method, treating all attention heads equally disregards their distinct roles, such as capturing contextual or positional information~\cite{jo2020roles}, leading to the dilution of critical insights. Equal treatment obscures meaningful patterns by averaging significant features with less relevant ones~\cite{chefer2021transformer}. Additionally, as attention layers are hierarchically connected, disregarding the importance of specific heads can distort the propagation of crucial information, undermining the model’s interpretability and hindering reliable explainability in Transformer architectures.

In the Multi-Head Attention mechanism, each head captures different aspects of input data, learning diverse features like local or global dependencies~\cite{voita2019analyzing}. Some heads focus on task-critical information, such as dependencies for object detection or sentiment analysis, while others may extract redundant or less useful patterns~\cite{li2023does}. Identifying the most influential heads provides valuable insights into how the model prioritizes information~\cite{jin2024moh}. As shown in Figure~\ref{fig:headbar}, attention head weights within a layer often exhibit notable differences in their contributions.

\subsection{Weighted GMAR}
\label{ssec:me_gmar}

To improve the interpretability of ViTs, we propose the Weighted GMAR method (Algorithm~\ref{alg:gmar}). This approach incorporates gradient-based head importance into the attention rollout process, refining the aggregation of multi-head self-attention layers. The method consists of two main steps: (1) computing gradient-based head weights $w_h$ and (2) using these weights to compute the attention rollout.

\begin{algorithm}
\caption{\texttt{Gradient-Driven Multi-Head Attention Rollout (GMAR)}. pre-trained model $\mathcal{M}$, input tensor $I$, model output $F$, gradients of each head $G_h$, head weights $w$, attention Layers $A$, residual ratio $\alpha$}
\label{alg:gmar}
\begin{algorithmic}
\STATE \textbf{Compute Gradient-Based Head Weights $w_h$:}
\bindent
    \STATE $\mathcal{M}.\text{zero\_grad()}$
    \STATE $F \gets \mathcal{M}(I)$
    \STATE $y_\text{pred} \gets max(F.\text{logits})$
    \STATE $\text{target} \gets F.\text{logits}[0, y_\text{pred}]$    
    \STATE $\text{target}.\text{backward()}$    
    \STATE $G \gets \text{Extract gradients of attention layers}$
    \STATE $G_h \gets split(G, num\_head)$ \COMMENT{Split by channel}    
    \FOR{each head index $i \in \{1, 2, \dots, num\_head\}$}        
        \STATE \small $G_R =
        \begin{cases} 
            \sum |G_{hi}| & \scriptsize \text{if \textbf{L1} regularization} \\
            \sqrt{\sum G_{hi}^2} & \scriptsize \text{if \textbf{L2} regularization}
        \end{cases}$        
    \ENDFOR
    \STATE \normalsize $w \gets \frac{G_R}{\sum G_R}$
\eindent
\STATE
\STATE \textbf{Compute Attention Rollout with Head Weights:}
\bindent
\STATE Initialize $A_{\text{rollout}} \gets \mathbb{I}_{N \times N}$ \COMMENT{Identity matrix for tokens}
\STATE $A_{\text{layers}} \gets F.\text{attentions}$
\FOR{each layer $A_{\ell} \in A_{\text{layers}}$}
    \STATE $W \gets w. \text{reshape}(1, H, 1, 1)$
    \STATE $A_{\text{weighted}} \gets A_{\ell} \cdot W$
    \STATE $A_{\text{rollout}} \gets A_{\text{rollout}} \cdot A_{\text{weighted}} + \alpha \cdot \mathbb{I}_{N \times N}$
\ENDFOR
\STATE \textbf{return} $A_{\text{rollout}}$
\eindent
\end{algorithmic}
\end{algorithm}
\vspace{-3mm}

\subsubsection{Gradient-Based Head Weights Computation}

Given a pre-trained ViT $\mathcal{M}$, we first compute the gradients of the predicted class logit $y_{\text{pred}}$ with respect to the attention layers. Specifically, for an input tensor $I$, the output logits $F = \mathcal{M}(I)$ are obtained, and the target logit $F_{\text{logits}}[0, y_{\text{pred}}]$ is backpropagated to compute gradients $G$. These gradients are split across the heads of each attention layer ($G_h$). To quantify the importance of each head, we apply a regularization-based weighting scheme. The gradient-based importance $G_R$ for head $i$ is defined as:

\begin{equation}
G_R = 
\begin{cases} 
\sum |G_{hi}|, & \text{if L1 regularization,} \\
\sqrt{\sum G_{hi}^2}, & \text{if L2 regularization.}
\end{cases}
\end{equation}

The weights $w$ are then normalized across all heads using the formula $w = \frac{G_R}{\sum G_R}$. This gradient-derived weight $w_h$ assigns higher importance to attention heads that exhibit stronger gradient magnitudes, signifying their contribution to the model’s decision.

\subsubsection{Weighted Attention Rollout}

Using the computed head weights, we enhance the attention rollout process to incorporate head-specific contributions. The rollout begins with an identity attention matrix $A_{\text{rollout}} \in \mathbb{R}^{N \times N}$, representing token-to-token relationships. For each attention layer $A_\ell$, the gradient-based weights $w$ are reshaped to match the dimensions of the attention heads and applied to the attention map is $A_{\text{weighted}} = A_\ell \cdot W$, where $W \in \mathbb{R}^{1 \times H \times 1 \times 1}$ represents the head weights. The weighted attention map $A_{\text{weighted}}$ is then recursively aggregated:
\begin{equation}
A_{\text{rollout}} = A_{\text{rollout}} \cdot A_{\text{weighted}} + \alpha \cdot \mathbb{I}_{N \times N}.
\end{equation}

Here, $\alpha$ controls the strength of the residual contribution, while $\mathbb{I}_{N \times N}$ is the identity matrix, ensuring that direct token-to-token relationships are retained alongside the aggregated attention. This iterative process integrates head importance into the multi-layer attention aggregation, producing a final attention rollout $A{\text{rollout}}$ that reflects both hierarchical, head-specific contributions, and residual connections that preserve essential information flow.

By leveraging gradient-based weights, GMAR captures the nuanced importance of individual attention heads, enhancing the interpretability of ViTs and improving the alignment between model decisions and input regions.

\section{Experiment}
\label{sec:experiment}

\subsection{Experimental Setup}
\label{ssec:ex_setup}

We utilize the Google ViT-Large-Patch16-224 model~\cite{wu2020visual, deng2009imagenet} as the base architecture, fine-tuning it on the Tiny-ImageNet dataset~\cite{le2015tiny}. This dataset was selected for its compact size, which enables rapid fine-tuning and experimentation. Although Tiny-ImageNet may yield slightly lower accuracy compared to larger datasets, it is well-suited for efficient iterative research. The fine-tuning process employs the Adam optimizer with $\beta_1 = 0.9$, $\beta_2 = 0.999$, and $\epsilon = 1 \times 10^{-8}$. The learning rate is set to $5 \times 10^{-5}$, and the training batch size is 32. The model is trained for 100 epochs, utilizing standard data augmentation techniques such as random cropping and flipping to improve generalization. All experiments are conducted on an NVIDIA RTX 2080 SUPER GPU.

\subsection{Dataset}
\label{ssec:ex_dataset}

The experiments in this study utilize the Tiny-ImageNet dataset, a widely recognized benchmark for evaluating image classification models. The dataset consists of 200 distinct classes, with each class containing 500 training images, 50 validation images, and an additional 50 test images. The images in Tiny-ImageNet are provided at a resolution of $64 \times 64$ pixels, which is smaller than the original ImageNet resolution but sufficient to capture meaningful patterns for classification tasks. For fine-tuning the ViT model, the images were resized to $224 \times 224$ pixels to match the input requirements of the \textit{ViT-Large-Patch16-224} architecture.

\subsection{Evaluation Metric}
\label{ssec:ex_metric}

To evaluate the effectiveness of the proposed explainability methods, we employ four widely used metrics in XAI evaluation~\cite{kadir2023evaluation}: Average Drop, Average Increase, Insertion, and Deletion. These metrics quantitatively assess the fidelity and relevance of the generated explanations.

Average Drop measures the average percentage decrease in confidence for the target class \( c \) when the model processes only the explanation map instead of the full image, defined as \( \left(\max(0, y_c - o_c) / y_c \right) \cdot 100 \), where \( y_c \) is the output score for the full image and \( o_c \) is the score for the explanation map, averaged over a set of images. Conversely, Average Increase calculates the percentage of cases where the confidence score is higher using the explanation map than the full image, expressed as \( \mathbf{I}_{y_c < o_c} \cdot 100 \), with \( \mathbf{I} \) being the indicator function, also averaged over multiple images. Additionally, Insertion and Deletion metrics evaluate model performance by measuring the rise and drop in the target class probability, respectively, as pixels are added or removed based on the explanation map, quantified as the total Area Under the Curve (AUC).

\begin{figure*}
    \centering
    \includegraphics[width=\textwidth]{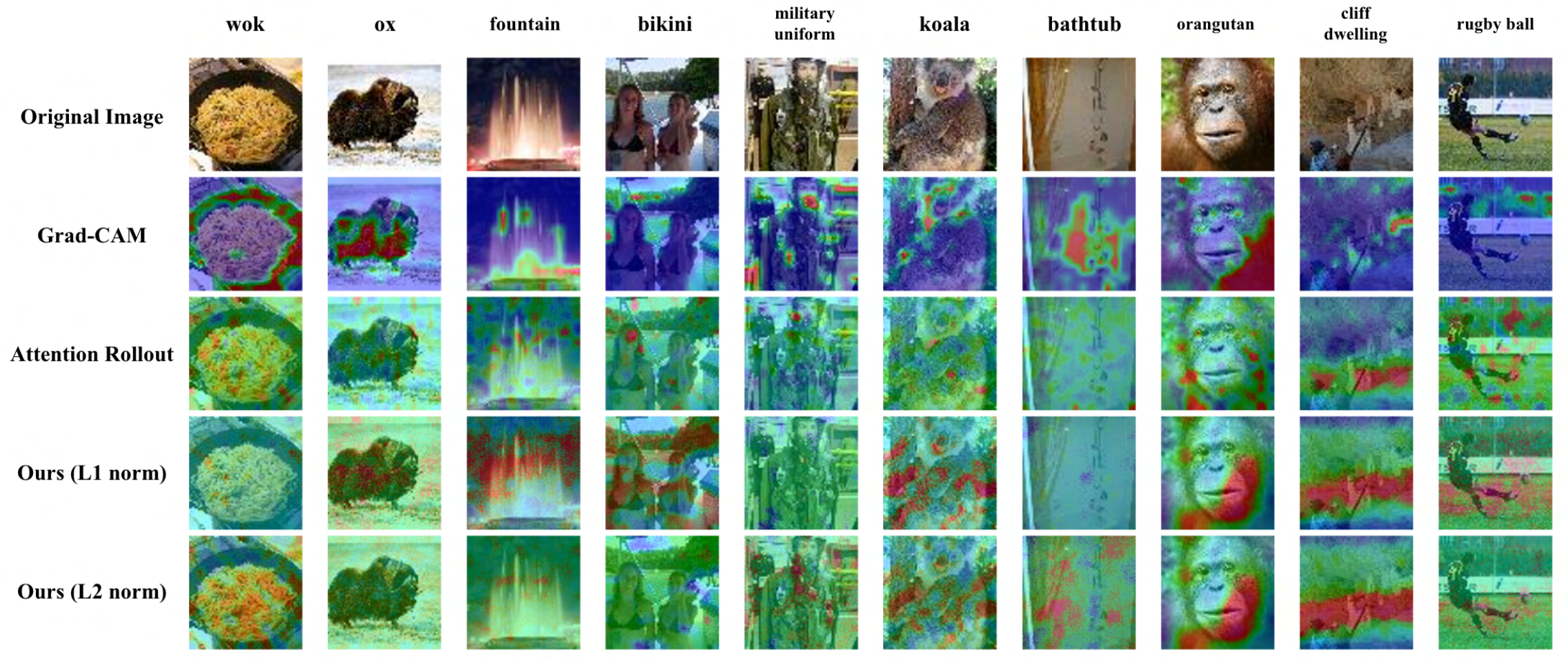}
    \caption{Overlay images generated for 10 randomly selected images from the Tiny ImageNet 200 dataset. The methods used include Grad-CAM, Attention Rollout, and GMAR with L1 and L2 norms, highlighting the regions of interest identified by each approach.}
    \label{fig:result}
\end{figure*}
\vspace{-3mm}

\section{Result}
\label{sec:result}

\subsection{Quantitative Results}

Table \ref{table:metric-table} summarizes the evaluation results of the interpretability methods for Vision Transformers. Grad-CAM outperforms Attention Rollout on most metrics, highlighting its effectiveness in identifying class-specific regions. However, Attention Rollout excels in the Insertion metric, indicating its strength in capturing broader contextual information.

Compared to Attention Rollout, GMAR outperforms across all evaluation metrics but shows slightly lower average increase and deletion scores than Grad-CAM due to their differing focus—CAM-based methods highlight localized discriminative regions, while attention-based approaches capture global context. Our findings confirm that GMAR enhances the interpretability of existing attention-based methods by effectively integrating gradient and attention signals. Notably, the observed performance differences between L1 and L2 norms indicate that GMAR not only identifies substantial predictive changes (L1 norm) when critical regions are altered but also accounts for gradual, cumulative variations (L2 norm), providing a more comprehensive interpretability framework. These results suggest that gradient-based approaches, which encode class-specific contributions, produce more faithful interpretability maps than attention-weight-based methods. GMAR’s integration of gradient and attention information further enhances visualization robustness and detail, strengthening its effectiveness in explaining model decisions.

\vspace{-2mm}
\begin{table}[h!]
    \centering
    \caption{Evaluation of interpretability methods using metrics such as Average Drop, Average Increase, Insertion, and Deletion.}
    \resizebox{\columnwidth}{!}{%
    \begin{tabular}{c|c|c|c|c}
        \hline
        \textbf{Method} & \textbf{Avg Drop} $\downarrow$ & \textbf{Avg Inc} $\uparrow$ & \textbf{Insertion} $\uparrow$ & \textbf{Deletion} $\downarrow$ \\ 
        \hline
        Grad-CAM & 22.61 & 65.8 & 10.75 & 10.09\\ 
        \hline
        Attention Rollout & 25.78 & 46.2 & 11.97 & 12.17\\ 
        GMAR (L1 norm) & 23.93 & \textbf{56.1} & 12.15 & \textbf{10.62} \\ 
        GMAR (L2 norm) & \textbf{22.13} & 55.9 & \textbf{12.16} & 10.64 \\ 
        \hline        
    \end{tabular}
    }
    \label{table:metric-table}
\end{table}
\vspace{-3mm}

\subsection{Qualitative Results}

The qualitative results clearly demonstrate the effectiveness of GMAR over standard attention rollout in highlighting objects of interest, as illustrated in Figure~\ref{fig:result}. While attention rollout often struggles to capture targeted objects accurately, GMAR consistently produces more focused and precise attention maps. Additionally, integrating Grad-CAM with attention rollout confirms the appropriate application of gradients, further enhancing the interpretability of these visualizations.

\begin{figure}
    \centering
    \includegraphics[width=\columnwidth]{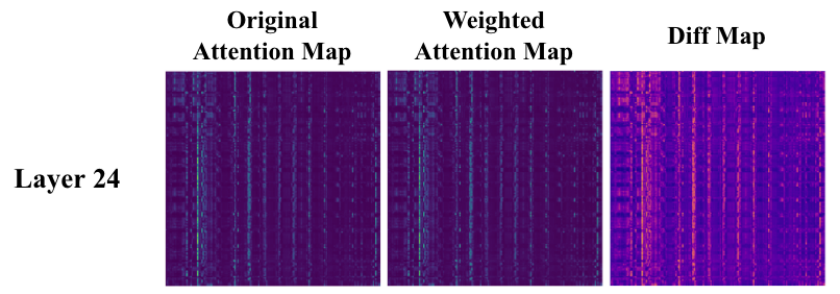}
    \caption{Visualization of the attention map, weighted attention map, and difference map.}
    \label{fig:attnmap}
\end{figure}

Moreover, the difference map analysis in Figure~\ref{fig:attnmap} indicates that weighted attention maps yield higher confidence scores than the original maps. Brighter regions highlight an intensified emphasis on essential areas, implying that the weighted attention mechanism not only boosts prediction confidence but also provides clearer insight into the model’s focus and reasoning. These findings highlight GMAR's ability to enhance attention visualizations, ensuring reliable and transparent model decisions.

\vspace{-3mm}
\vspace{-3mm}

\section{Conclusion}
\label{sec:result}

Our paper introduced GMAR, a novel framework to enhance the interpretability of ViTs by assigning gradient-based importance scores to individual attention heads, enabling a more precise understanding of their contributions to model predictions. GMAR improves upon attention rollout by incorporating these scores into the process, resulting in attention maps that more accurately reflect input relevance and model decisions. Experimental results demonstrated GMAR's superiority in key interpretability metrics confirming its ability to identify task-critical regions and maintain prediction confidence. Future work will explore the scalability of GMAR by applying it to larger datasets and advanced ViT architectures, ensuring its effectiveness in diverse applications.

\bibliographystyle{IEEEbib}
\bibliography{strings,refs}

\begin{thebibliography}{10}

\bibitem{dosovitskiy2020image}
Alexey Dosovitskiy,
\newblock ``An image is worth 16x16 words: Transformers for image recognition at scale,''
\newblock {\em arXiv preprint arXiv:2010.11929}, 2020.

\bibitem{ali2022xai}
Ameen Ali, Thomas Schnake, Oliver Eberle, Gr{\'e}goire Montavon, Klaus-Robert M{\"u}ller, and Lior Wolf,
\newblock ``Xai for transformers: Better explanations through conservative propagation,''
\newblock in {\em ICML}, 2022.

\bibitem{abnar2020quantifying}
Samira Abnar and Willem Zuidema,
\newblock ``Quantifying attention flow in transformers,''
\newblock in {\em ACL}, 2020.

\bibitem{ribeiro2016should}
Marco~Tulio Ribeiro, Sameer Singh, and Carlos Guestrin,
\newblock ``" why should i trust you?" explaining the predictions of any classifier,''
\newblock in {\em SIGKDD}, 2016.

\bibitem{scott2017unified}
M~Scott, Lee Su-In, et~al.,
\newblock ``A unified approach to interpreting model predictions,''
\newblock {\em NeurIPS}, 2017.

\bibitem{selvaraju2017grad}
Ramprasaath~R Selvaraju, Michael Cogswell, Abhishek Das, Ramakrishna Vedantam, Devi Parikh, and Dhruv Batra,
\newblock ``Grad-cam: Visual explanations from deep networks via gradient-based localization,''
\newblock in {\em ICCV}, 2017.

\bibitem{bach2015pixel}
Sebastian Bach, Alexander Binder, Gr{\'e}goire Montavon, Frederick Klauschen, Klaus-Robert M{\"u}ller, and Wojciech Samek,
\newblock ``On pixel-wise explanations for non-linear classifier decisions by layer-wise relevance propagation,''
\newblock {\em PloS one}, vol. 10, no. 7, pp. e0130140, 2015.

\bibitem{xie2023vit}
Weiyan Xie, Xiao-Hui Li, Caleb~Chen Cao, and Nevin~L Zhang,
\newblock ``Vit-cx: causal explanation of vision transformers,''
\newblock in {\em IJCAI}, 2023.

\bibitem{panda2024human}
Mahadev~Prasad Panda, Matteo Tiezzi, Martina Vilas, Gemma Roig, Bjoern~M Eskofier, and Dario Zanca,
\newblock ``Human-inspired explanations for vision transformers and convolutional neural networks,''
\newblock {\em arXiv preprint arXiv:2408.02123}, 2024.

\bibitem{vaswani2017attention}
A~Vaswani,
\newblock ``Attention is all you need,''
\newblock {\em NeurIPS}, 2017.

\bibitem{devlin2018bert}
Jacob Devlin,
\newblock ``Bert: Pre-training of deep bidirectional transformers for language understanding,''
\newblock {\em arXiv preprint arXiv:1810.04805}, 2018.

\bibitem{liu2019roberta}
Yinhan Liu,
\newblock ``Roberta: A robustly optimized bert pretraining approach,''
\newblock {\em arXiv preprint arXiv:1907.11692}, vol. 364, 2019.

\bibitem{brown2020language}
Tom Brown, Benjamin Mann, Nick Ryder, Melanie Subbiah, Jared~D Kaplan, Prafulla Dhariwal, Arvind Neelakantan, Pranav Shyam, Girish Sastry, Amanda Askell, et~al.,
\newblock ``Language models are few-shot learners,''
\newblock {\em NeurIPS}, 2020.

\bibitem{touvron2023llama}
Hugo Touvron, Thibaut Lavril, Gautier Izacard, Xavier Martinet, Marie-Anne Lachaux, Timoth{\'e}e Lacroix, Baptiste Rozi{\`e}re, Naman Goyal, Eric Hambro, Faisal Azhar, et~al.,
\newblock ``Llama: Open and efficient foundation language models,''
\newblock {\em arXiv preprint arXiv:2302.13971}, 2023.

\bibitem{touvron2021training}
Hugo Touvron, Matthieu Cord, Matthijs Douze, Francisco Massa, Alexandre Sablayrolles, and Herv{\'e} J{\'e}gou,
\newblock ``Training data-efficient image transformers \& distillation through attention,''
\newblock in {\em ICML}, 2021.

\bibitem{bao2021beit}
Hangbo Bao, Li~Dong, Songhao Piao, and Furu Wei,
\newblock ``Beit: Bert pre-training of image transformers,''
\newblock {\em arXiv preprint arXiv:2106.08254}, 2021.

\bibitem{liu2021swin}
Ze~Liu, Yutong Lin, Yue Cao, Han Hu, Yixuan Wei, Zheng Zhang, Stephen Lin, and Baining Guo,
\newblock ``Swin transformer: Hierarchical vision transformer using shifted windows,''
\newblock in {\em CVPR}, 2021.

\bibitem{jo2020roles}
Jae-young Jo and Sung-Hyon Myaeng,
\newblock ``Roles and utilization of attention heads in transformer-based neural language models,''
\newblock in {\em ACL}, 2020.

\bibitem{chefer2021transformer}
Hila Chefer, Shir Gur, and Lior Wolf,
\newblock ``Transformer interpretability beyond attention visualization,''
\newblock in {\em CVPR}, 2021.

\bibitem{voita2019analyzing}
Elena Voita, David Talbot, Fedor Moiseev, Rico Sennrich, and Ivan Titov,
\newblock ``Analyzing multi-head self-attention: Specialized heads do the heavy lifting, the rest can be pruned,''
\newblock {\em arXiv preprint arXiv:1905.09418}, 2019.

\bibitem{li2023does}
Yiran Li, Junpeng Wang, Xin Dai, Liang Wang, Chin-Chia~Michael Yeh, Yan Zheng, Wei Zhang, and Kwan-Liu Ma,
\newblock ``How does attention work in vision transformers? a visual analytics attempt,''
\newblock {\em IEEE transactions on visualization and computer graphics}, 2023.

\bibitem{jin2024moh}
Peng Jin, Bo~Zhu, Li~Yuan, and Shuicheng Yan,
\newblock ``Moh: Multi-head attention as mixture-of-head attention,''
\newblock {\em arXiv preprint arXiv:2410.11842}, 2024.

\bibitem{wu2020visual}
Bichen Wu, Chenfeng Xu, Xiaoliang Dai, Alvin Wan, Peizhao Zhang, Zhicheng Yan, Masayoshi Tomizuka, Joseph Gonzalez, Kurt Keutzer, and Peter Vajda,
\newblock ``Visual transformers: Token-based image representation and processing for computer vision,'' 2020.

\bibitem{deng2009imagenet}
Jia Deng, Wei Dong, Richard Socher, Li-Jia Li, Kai Li, and Li~Fei-Fei,
\newblock ``Imagenet: A large-scale hierarchical image database,''
\newblock in {\em CVPR}, 2009.

\bibitem{le2015tiny}
Yann Le and Xuan Yang,
\newblock ``Tiny imagenet visual recognition challenge,''
\newblock {\em CS 231N}, vol. 7, no. 7, pp. 3, 2015.

\bibitem{kadir2023evaluation}
Md~Abdul Kadir, Amir Mosavi, and Daniel Sonntag,
\newblock ``Evaluation metrics for xai: A review, taxonomy, and practical applications,''
\newblock in {\em IEEE INES}, 2023.

\end{thebibliography}

\end{document}